\documentclass{article}

\PassOptionsToPackage{numbers, sort&compress}{natbib}

\usepackage{multirow}

\usepackage[preprint]{neurips_2026}
\usepackage{amsmath}

\usepackage{wrapfig}
\usepackage[table]{xcolor}
\usepackage[utf8]{inputenc} 
\usepackage[T1]{fontenc}    
\usepackage{hyperref}       
\usepackage{url}            
\usepackage{booktabs}       
\usepackage{amsfonts}       
\usepackage{nicefrac}       
\usepackage{microtype}      
\usepackage{xcolor}       
\usepackage{enumitem}
\usepackage{algorithm}
\usepackage{algpseudocode}
\usepackage{graphicx}
\usepackage{subcaption}
\usepackage{makecell}

\usepackage{amsmath,amssymb,amsthm}
\newtheorem{proposition}{Proposition}

\title{UniSteer: Unified Noise Steering for Efficient Human-Guided VLA Adaptation}

\author {
    \textbf{Junjie Lu}\textsuperscript{\rm 1}\footnotemark[1]\;\footnotemark[2] , \hspace{0.1cm}
    \textbf{Xinyao Qin}\textsuperscript{\rm 2}\footnotemark[1]\;\footnotemark[2] , \hspace{0.1cm}
    \textbf{Yuhua Jiang}\textsuperscript{\rm 2}, \hspace{0.1cm}
    \textbf{Kaixin Wang}\textsuperscript{\rm 3}, \hspace{0.1cm}
    \textbf{Chuheng Zhang}\textsuperscript{\rm 3}, \hspace{0.1cm}
    \\
    \textbf{Bin Liang}\textsuperscript{\rm 2}\textbf{,} \hspace{0.1cm}
    \textbf{Jun Yang}\textsuperscript{\rm 2}\textbf{,} \hspace{0.1cm}
    \textbf{Min Xu}\textsuperscript{\rm 1}\textbf{,} \hspace{0.1cm}
    \textbf{Li Zhao}\textsuperscript{\rm 3}\hspace{0.1cm} \\
    \textsuperscript{\rm 1} University of Technology Sydney
    \textsuperscript{\rm 2} Tsinghua University 
    \textsuperscript{\rm 3} Microsoft Research
}

\begin{document}
\renewcommand{\thefootnote}{\fnsymbol{footnote}}
\maketitle

\footnotetext[1]{Equal contribution.}
\footnotetext[2]{Work conducted at Microsoft Research.}

\renewcommand{\thefootnote}{\arabic{footnote}}
\setcounter{footnote}{0}

\begin{abstract}
Diffusion-based vision-language-action (VLA) models have emerged as strong priors for robotic manipulation, yet adapting them to real-world distributions remains challenging. In particular, on-robot reinforcement learning (RL) is expensive and time-consuming, so effective adaptation depends on efficient policy improvement within a limited budget of real-world interactions. Noise-space RL lowers the cost by keeping the pretrained VLA fixed as a denoising generator while updating only a lightweight actor that predicts the noise. However, its performance is still limited due to inefficient autonomous exploration. Human corrective interventions can reduce this exploration burden, but they are naturally provided in action space, whereas noise-space finetuning requires supervision over noise variables. To address these challenges, we propose \textbf{UniSteer}, a \textbf{Uni}fied Noise \textbf{Steer}ing framework that combines human corrective guidance with noise-space RL through approximate action-to-noise inversion. Given a human corrective action, UniSteer inverts the frozen flow-matching decoder to recover a noise target, which provides supervised guidance for the same noise actor that is simultaneously optimized via reinforcement learning. Real-world experiments on diverse manipulation tasks show that UniSteer adapts more efficiently than strong noise-space RL and action-space human-in-the-loop baselines, improving the success rate from 20\% to 90\% in 66 minutes on average across four real-world adaptation tasks.
\end{abstract}

\section{Introduction}
Vision-language-action (VLA) models have recently shown strong capability across diverse robotic tasks, including single-arm manipulation~\cite{brohan2022rt,zitkovich2023rt,zhu2023vima,kim2024openvla,team2024octo,zheng2025universal,zhao2025vlas,chen2025villa,upvla,univla,bridgevla, smolvla}, bimanual manipulation~\cite{o2024rtx,black2410pi0,pi05,bu2025agibot,liu2024rdt,kim2025fine,xvla} and dexterous manipulation~\cite{dexgraspvla,dexvla,zhang2026unidex, jiang2026cross}.
Among them, diffusion-based VLAs~\citep[e.g.,][]{black2410pi0,liu2024rdt,smolvla} become popular due to their strong performance.
However, when deployed in the real world, these pretrained models often encounter distribution shifts in scenes, objects, viewpoints, and contact dynamics. They can significantly degrade performance, especially in long-horizon, contact-rich, or high-precision manipulation tasks.
While reinforcement learning (RL) provides a natural way to adapt VLA policies through task-specific interactions \cite{conrft, dsrl, pi06, riptvla, vla-rl, reinbot, rlinf, ire-vla, simplevla, gr-rl, dualactor, srpo}, on-robot RL remains costly: collecting real-world experience is expensive, failures consume substantial resources, and adaptation is constrained by tight wall-clock time budgets.
This motivates the need for efficient VLA adaptation methods under limited on-robot interaction and wall-clock time budget.

\begin{figure}[t]
  \centering
  \includegraphics[width=0.85\linewidth,]{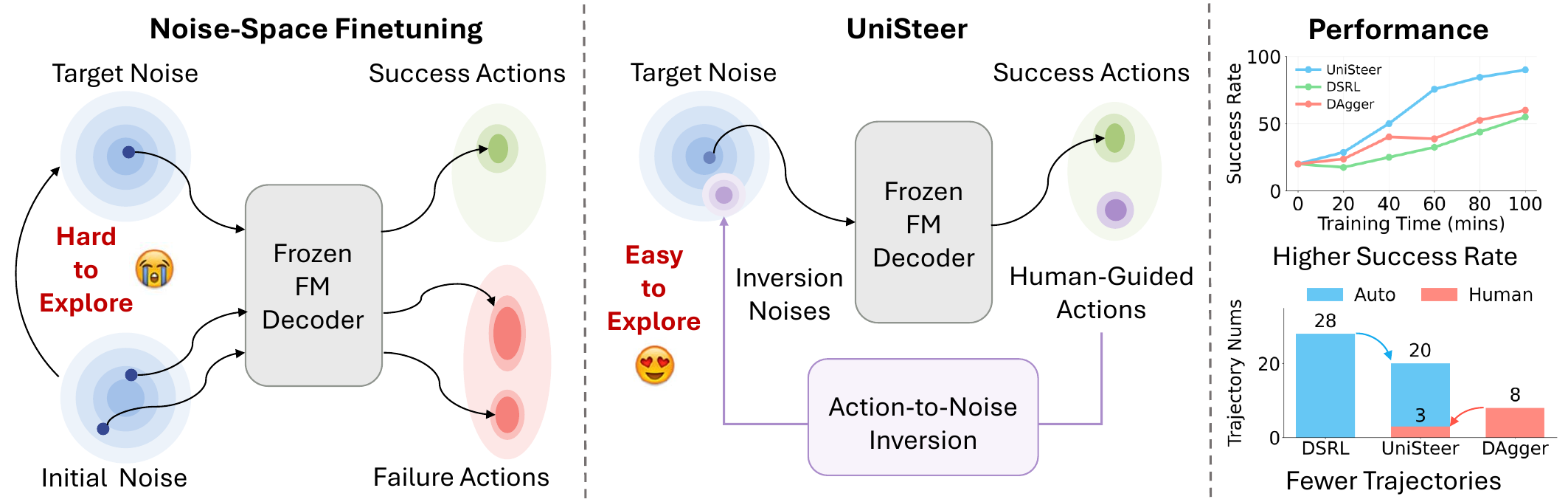}
  \caption{\textbf{UniSteer bridges human guidance and noise-space RL.} Noise-space finetuning relies on exploration from a distant initial noise distribution. UniSteer maps human corrective actions into noise-space targets, providing a useful prior for faster adaptation with fewer trajectories.}
  \vspace{-0.3cm}
\end{figure}

One efficient way to avoid the high computational costs of adapting diffusion-based generative policies is \textit{noise-space finetuning} \cite{dsrl, golden-ticket, gr-rl,dualactor}. Specifically, we freeze the pretrained diffusion-based VLA as a denoising generator and steer a small noise actor to generate the noise.
This confines adaptation to a small learnable module, reducing optimization cost while keeping the pretrained VLA frozen. As a result, the decoded actions naturally remain anchored to the pretrained policy prior, thus benefit from the priors in pretrained VLAs.
However, since the VLA can degenerate under out-of-distribution (OOD) deployment conditions, policy improvement depends heavily on autonomous exploration, which can be inefficient under sparse rewards or large deployment shifts.

A natural way to mitigate this exploration bottleneck is to incorporate human guidance. While existing approaches for demonstrations~\cite{dagger}, corrective interventions~\cite{hg-dagger, hu2025rac, human-reject}, and human-in-the-loop learning~\cite{hil-serl,conrft,dualactor,apohuman} provide effective ways to refine robot behavior, they typically operate within the \textit{action space}. 
Such action-space updates are often heavier and less stable compared with \textit{noise-space finetuning}, since they require optimizing the action-generation pathway of a large generative policy.
However, incorporating human corrective actions into noise-space is non-trivial. The mapping from noise to action is induced by a complex, learned transport process. 
Consequently, recovering a noise variable from a human-corrected action requires an implicit reverse Euler update
whose velocity depends on the unknown preimage\cite{lai2025principles,lipman2022flow}. Alternatively, backpropagating action-level supervision through the entire decoding chain is another option but it is computationally prohibitive and prone to unstable optimization signals.

To address this challenge, we propose \textbf{UniSteer}, a \textbf{Uni}fied Noise \textbf{Steer}ing framework that integrates human corrective guidance into noise-space online RL through approximate action-to-noise inversion. 
Although this inversion is not solved directly in closed form, we show that a tractable approximation yields low inversion error and provides reliable supervision in practice. 
Building on this inversion, we introduce a noise-space finetuning framework that unifies RL and human guidance through a shared noise-space interface, allowing 
the use of human corrections to supervise the noise actor that optimized via RL in the meanwhile.
We validate this framework through extensive real-world robotic adaptation experiments, 
showing that it outperforms strong baselines with reduced time-to-performance and fewer human interventions. Our contributions are threefold: 
\begin{enumerate}[label=(\arabic*), leftmargin=*]
\item We introduce UniSteer, the first unified noise-space steering framework that incorporates human guidance into noise-space online RL for efficient real-world adaptation.
\item We propose an approximate action-to-noise inversion method for flow-matching policies, enabling consistent noise supervision under a frozen generator.
\item We demonstrate that our method achieves reduced time-to-performance with fewer human interventions than strong baselines across diverse real-world robotic manipulation tasks.
\end{enumerate}

\section{Related work}

\subsection{Vision-Language-Action Models}
Vision-Language-Action (VLA) models have become a central paradigm for robot learning~\cite{brohan2022rt,zitkovich2023rt,zhu2023vima,kim2024openvla,team2024octo,zheng2025universal,zhao2025vlas,chen2025villa,upvla,univla,bridgevla, smolvla, o2024rtx,black2410pi0,pi05,bu2025agibot,liu2024rdt,kim2025fine,xvla}.
Early methods formulate action prediction as autoregressive token generation to integrate robot control into language-model-style sequence modeling~\cite{brohan2022rt,zitkovich2023rt,zhu2023vima,o2024rtx,kim2024openvla,kim2025fine,univla}.
Other methods attach continuous action heads or diffusion-style decoders to vision-language models for better representation of high-dimensional continuous control~\cite{team2024octo,liu2024rdt,hou2025dita,wen2024diffusionvla}.
More recently, many VLA policies adopt flow-matching action heads, showing strong generative capability and promising performance in real-world robotic manipulation~\cite{black2410pi0,pi05,bu2025agibot,chen2025villa,smolvla,xvla}.
These policies generate action chunks by transporting initial noise variables to continuous actions through a learned state-conditioned velocity field.
Our work focuses on this class of flow-matching VLA policies due to their competitive performance and leverages the initial noise variable as a natural interface for policy steering.

\subsection{Adaptation of Flow-Matching VLA Policies}

Reinforcement learning provides a natural approach for downstream VLA adaptation, but directly applying it to flow-matching policies is challenging.
These policies generate actions through multi-step decoding, making end-to-end optimization costly and unstable. Moreover, their deterministic generation process makes the action log probabilities or likelihood ratios required by many RL algorithms difficult to obtain.
Recent methods address this challenge by reformulating policy-gradient objectives for flow policies, such as RLinf~\cite{rlinf} and FPO~\cite{mcallister2025flow}.
Other works avoid difficult online RL updates in the real world and instead improve policies through demonstrations, human interventions, or value-weighted behavior cloning, including HG-DAgger~\cite{hg-dagger}, AWR-style methods~\cite{su2026ig,yang2026aloe,mao2026arm}, and RECAP-style value-weighted imitation~\cite{pi06}.
Another line of work keeps the pretrained generative policy frozen and adapts only a lightweight module, such as a residual correction network\cite{xiao2025self,xu2026rltokenbootstrappingonline} or a noise-space policy~\cite{dsrl,golden-ticket,gr-rl,dualactor}. Compared with the above two lines of work, this paradigm offers a more lightweight and stable adaptation strategy by freezing the pretrained VLA and updating only a small module.
Our method follows this lightweight adaptation paradigm, but differs from prior noise-space methods that mainly rely on autonomous exploration: we translate action-space human corrections into noise-space supervision targets, allowing human interventions to directly guide the compact noise actor while keeping the flow-matching VLA frozen.

\section{Problem Formulation}

\subsection{Flow matching policies.}
We consider VLA policies parameterized as conditional flow matching models. 
Given a state $s$, the policy maps an initial noise variable $z_0 \sim \mathcal{N}(0,I)\in \mathbb{R}^d$ to an action chunk $a \in \mathbb{R}^d$ by integrating a state-conditioned velocity field $v_\theta(z_t,t,s)$, where $\theta$ is the parameter of the network:
\[
\frac{d z_t}{dt} = v_\theta(z_t,t,s).
\]
At each flow step $k$, the network predicts a velocity update conditioned on the current noise, the flow time, and the state. 
Starting from $z_0$ at $t=0$ and integrating forward to $t=1$ produces the terminal sample deterministically
\[
a = G_\theta(s, z_0) := z_1 = z_0 + \int_{0}^{1} v_\theta(z_t,t,s)\,dt,
\]
where $a$ denotes the predicted action chunk. In practice, this continuous-time process is implemented using a K-step Euler discretization with step size $\Delta t$. 

\subsection{Noise-space policy optimization.}
We formulate robotic policy learning as a Markov Decision Process (MDP)
$\mathcal{M}=(\mathcal{S},\mathcal{A},P,r,\gamma)$, where $\mathcal{S}$ is the state space, 
$\mathcal{A}\subseteq\mathbb{R}^d$ is the continuous action space, 
$P(s' \mid s,a)$ is the transition dynamics, 
$r(s,a)$ is the reward function, 
and $\gamma \in [0,1)$ is the discount factor. In our setting, the action is in fact the noise variable $z$ because we freeze the decoding process $G_\theta(s,z)$.
Specially, a lightweight policy $\psi_\phi(z \mid s)$ is trained to select the initial noise variable conditioned on the current state, and the frozen decoder maps it to an action chunk:
\[
z \sim \psi_\phi(\cdot \mid s), \qquad a = G_\theta(s,z).
\]
We further learn a noise-space critic $Q_\omega(s,z)$, which estimates the expected discounted return of the trajectory. $Q_\omega(s,z)$ is trained with the standard TD loss:
\[
\mathcal{L}_{Q_\omega} = 
\left\|
Q_\omega(s,z) - \big(r + \gamma \bar Q_\omega(s',z')\big)
\right\|^2,
\]

The RL objective is therefore to optimize the policy with respect to this value function, while the pretrained flow decoder remains fixed.
\[
\mathcal{L}_{\mathrm{RL}}
=
-
Q_\omega(s,z),
\qquad z \sim \psi_\phi(s).
\]

\section{Method}

\subsection{Action-to-Noise Inversion}\label{4.1}

Given the current state $s$ and a human-provided corrective action chunk $a^h$, our goal is to recover a noise target $\hat z$ such that, the frozen flow-matching policy can decode it into an action chunk close to $a^h$. Ideally, the desired noise target is the inverse image $G_\theta(s,\cdot)^{-1}(a^h)$. However, since this noise-to-action mapping is realized through a multi-step nonlinear flow decoder, directly inverting it is difficult. 
We therefore exploit its stepwise Euler structure and invert the decoder one step at a time along the frozen decoding trajectory.

\begin{figure}[t]
  \centering
  \includegraphics[width=\linewidth]{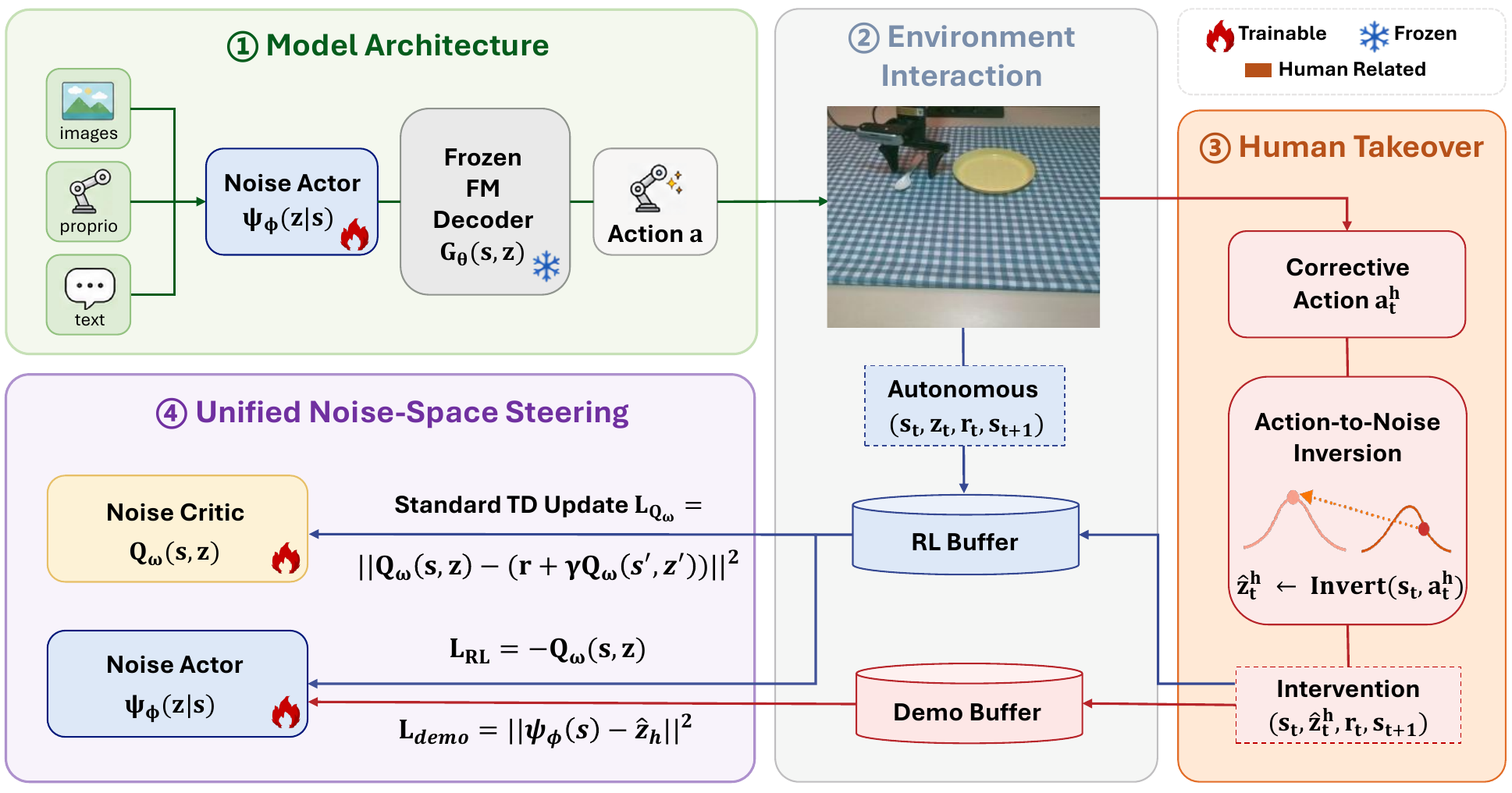}
\caption{\textbf{Overview of UniSteer framework.} The noise actor maps current states into noise variables and generates actions through the frozen decoder. Autonomous rollouts are stored in the RL buffer, whereas human takeover actions are inverted into noise space and collected in both the demo buffer and the RL buffer. The two buffers jointly provide training signals for the critic and the actor.}
\end{figure}

\begin{proposition}
Suppose $v_\theta(z,t,s)$ is continuous in $t$ and globally Lipschitz in $z$,
uniformly for $t \in [0,1]$\footnote{For each fixed state $s$, assume that $t \mapsto v_\theta(z,t,s)$ is continuous
for every $z \in \mathbb{R}^d$, and that there exists a constant $L_s > 0$
such that $\|v_\theta(z,t,s)-v_\theta(z',t,s)\| \le L_s \|z-z'\|,
\forall z,z' \in \mathbb{R}^d,\ \forall t \in [0,1].
$}.
Then for a fixed state $s$, the map $G_\theta(s,\cdot):\mathbb{R}^d \to \mathbb{R}^d$
is bijective. 
In particular, for every action $a \in \mathbb{R}^d$, there exists a unique
$z_0 \in \mathbb{R}^d$ such that $a = G_\theta(s,z_0)$.
\end{proposition}

\textbf{Remark.}
This proposition establishes the controllability of the noise-space actor.
Intuitively, the global Lipschitz continuity of the velocity field prevents constant mapping or squashed mapping (such as $G_\theta(s, z_0) = \tanh(z_0)$).

For the frozen flow matching policy, forward decoding proceeds through  $K$ iterative Euler updates. At flow step $k$, the inverse of a single update can be written as a fixed-point equation:
\[
x = y - \Delta t\, v_\theta(x,t_k,s) = g_y(x).
\]
where $x$ and $y$ denote the input and output of that decoder step, respectively, and $v_\theta$ is the frozen velocity model. 

\begin{proposition}
Assume that $v_\theta(\cdot,t_k,s)$ is $L$-Lipschitz. If $\Delta t L < 1$, then the map
$g_y(x)=y-\Delta t\,v_\theta(x,t_k,s)$ is contractive. 
\end{proposition}

Consequently, the inversion of the Euler step exists uniquely and can be recovered by fixed-point iteration. 
Starting from the corrective action chunk as the terminal state, we recover the noise states recursively as
\[
\hat z_k = \mathrm{Inv}_k(\hat z_{k-1}, s), \qquad k=1,\dots,K, \qquad \hat z_0 = a^h
\]
\[
\mathrm{Inv}_k(\hat z_{k-1}, s)
\;:=\;
\operatorname*{fix}\Big(z \mapsto \hat z_{k-1} - \Delta t\, v_\theta(z,t_k,s)\Big).
\]
$\mathrm{Inv}_k$ is defined as the fixed point of the inverse Euler map.
In practice, we approximate this fixed point with $M$ steps of iteration:
\[
z_k^{(m+1)} = \hat z_{k-1} - \Delta t\, v_\theta\!\left(z_k^{(m)}, t_k, s\right), \quad m=0,\dots,M-1, \qquad z_k^{(0)}=\hat z_{k-1}
\] 
After all $K$ backward steps, we obtain the recovered initial noise
$\hat z := \hat z_K,$
which is used as the noise supervision target.

\subsection{Unified Noise Steering Framework}
With the inversion procedure, human corrective actions can be translated into noise-space supervision targets under the frozen flow matching policy. During online interaction, the noise actor $\psi_\phi$ selects an initial noise variable conditioned on the current state, and the frozen flow matching policy decodes it into an action chunk:
\[
z_t \sim \psi_\phi(\cdot \mid s_t), \qquad a_t = G_\theta(s_t, z_t).
\]
During autonomous interaction, we execute the decoded action chunk $a_t$ and store the resulting transition $(s_t, z_t, r_t, s_{t+1})$ in the RL buffer $\mathcal{B}_{\mathrm{RL}}$. When a human takes over and provides a corrective action chunk $a_t^h$, we apply the inversion procedure in Sec\ref{4.1} to recover the corresponding noise target $\hat z_t^h$ and execute the corrective action. The corrected transition $(s_t, \hat z_t^h, r_t, s_{t+1})$ is stored in the RL buffer $\mathcal{B}_{\mathrm{RL}}$ for value learning and in the demo buffer $\mathcal{B}_{\mathrm{demo}}$ for correction supervision. This design allows human interventions to improve both critic learning and actor optimization under a unified noise-space representation.

For policy optimization, we first update the policy by sampling batches from demo buffer $\mathcal{B}_{\mathrm{demo}}$ by
\[
\mathcal{L}_{\mathrm{demo}}
= ||\psi_{\phi}(s) - \hat{z}_h||^2_2.
\]
Then we sample from $\mathcal{B}_{\mathrm{RL}}$ and conduct optimization using $\mathcal{L}_{\mathrm{RL}}$.
This design allows RL and human guidance to update the same noise actor through a shared noise interface. The RL branch improves the policy from reward feedback, while the correction branch injects action-space human guidance after translating it into noise-space targets through inversion. Algorithm~\ref{alg:latent_human_rl_short} summarizes the overall training procedure.

\section{Experiment}

\subsection{Experimental Setup}

\begin{figure}[b]
  \centering
  \includegraphics[width=\linewidth]{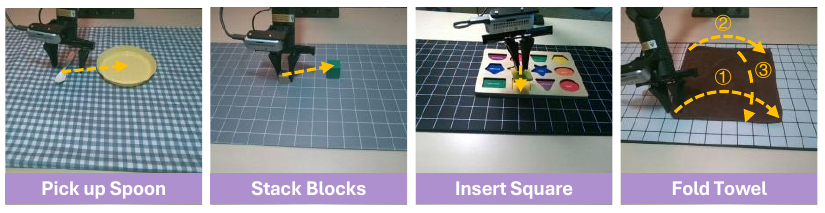}
  \caption{Overview of all real-world experimental tasks.}
  \label{fig:tasks}
\end{figure}

\paragraph{Tasks and Platform} We evaluate our method on four real-world manipulation tasks: \textit{pick up spoon}, \textit{stack blocks}, \textit{insert square}, and \textit{fold towel}, as shown in Figure~\ref{fig:tasks}. These tasks cover a broad range of manipulation skills, including pick-and-place, precise contact-rich operations, and deformable-object manipulation. Together, they provide a compact benchmark for evaluating both broad spatial generalization and fine-grained manipulation refinement under distribution shift. All experiments are conducted on an AgileX Piper robot with a master--slave teleoperation setup for human takeover. For all tasks, the observations consist of two RGB images from a side camera and a wrist camera, together with the robot's 6D end-effector pose and gripper state. The action space is defined as the end-effector target pose with gripper openness, and the robot is controlled at 30 Hz. We initialize the policy from a pre-trained $\pi_0$ architecture checkpoint with a flow-matching action head and warm it up using 30 task demonstrations per task. All training runs are conducted on a single NVIDIA A100 GPU.

\paragraph{Baselines}
To fairly evaluate the effectiveness of our method, we compare against two representative baselines: DSRL\cite{dsrl}, DAgger\cite{dagger}. DSRL serves as the primary noise-space RL baseline, as it performs policy improvement under the same frozen generative policy backbone but without human intervention, thereby isolating the benefit of incorporating corrective guidance. DAgger represents a human-in-the-loop imitation learning approach, where corrective actions are aggregated and used for supervised policy updates in action space. Together, these baselines cover the relevant comparison axes for our setting: noise-space RL without human assistance, action-space imitation learning with human corrections. We initially considered HIL-SERL\cite{hil-serl} as an additional human-in-the-loop online RL baseline. However, consistent with observations in RLToken~\cite{xu2026rltokenbootstrappingonline}, we found that HIL-SERL is poorly matched to our high-frequency control setting with sparse terminal rewards. In preliminary experiments, HIL-SERL was ineffective and unstable, and therefore did not provide a reliable comparison for our setting.

\paragraph{Evaluation Metrics}
We use real-world task success rate as the metric for evaluating each method's performance. For each task, the final policy is evaluated over 20 real-world trials, and the success rate is computed as the percentage of successful trials. For Pick up Spoon, Stack Blocks, and Insert Square, we randomly select 10 distinct initial object positions and evaluate each position twice. Among these positions, 20\% are out-of-distribution cases not covered by the initial demonstration data, allowing us to evaluate both in-distribution adaptation and spatial generalization to unseen object placements. For Fold Towel, since the task involves deformable-object manipulation and does not admit the same discrete position-based evaluation protocol, we directly run the policy for 20 consecutive trials and report the resulting success rate.

\subsection{Main Results}

\begin{table}[t]
\centering
\footnotesize
\setlength{\tabcolsep}{2.8pt}
\setlength{\belowcaptionskip}{6pt} 
\renewcommand{\arraystretch}{1.08}
\caption{Success rate (\%) after task-specific real-world adaptation. For position-based tasks, Overall is evaluated over 20 trajectories, while ID and OOD are evaluated over 16 in-distribution and 4 out-of-distribution trajectories, respectively. Numbers in parentheses denote absolute success-rate improvement over the initial policy for the same split. The Average row is computed across the four task-level Overall success rates.}
\label{tab:main_results}
\vspace{4pt}
\begin{tabular}{lccrrrr}
\toprule
\multirow{2}{*}{Task} 
& \multirow{2}{*}{\shortstack{Training\\Time (mins)}} 
& \multirow{2}{*}{Split} 
& \multicolumn{4}{c}{Success Rate (\%)} \\
\cmidrule(lr){4-7}
& & & \multicolumn{1}{c}{Initial Policy} & \multicolumn{1}{c}{UniSteer} & \multicolumn{1}{c}{DSRL} & \multicolumn{1}{c}{DAgger} \\
\midrule

\multirow{3}{*}{Pick up Spoon}
& \multirow{3}{*}{45} 
& ID  & 25.0\% 
& \textbf{87.5\% (+62.5\%)} 
& 62.5\% (+37.5\%) 
& 68.8\% (+43.8\%) \\
&                          
& OOD & 0.0\%  
& \textbf{100.0\% (+100.0\%)} 
& 0.0\% (+0.0\%) 
& 75.0\% (+75.0\%) \\
&
& Overall & 20.0\% 
& \textbf{90.0\% (+70.0\%)} 
& 50.0\% (+30.0\%) 
& 70.0\% (+50.0\%) \\

\midrule
\multirow{3}{*}{Stack Blocks}
& \multirow{3}{*}{60} 
& ID  & 43.8\% 
& \textbf{93.8\% (+50.0\%)} 
& 75.0\% (+31.3\%) 
& 62.5\% (+18.8\%) \\
&                          
& OOD & 0.0\%  
& \textbf{100.0\% (+100.0\%)} 
& 0.0\% (+0.0\%) 
& \textbf{100.0\% (+100.0\%)} \\
&
& Overall & 35.0\%
& \textbf{95.0\% (+60.0\%)}
& 60.0\% (+25.0\%)
& 70.0\% (+35.0\%) \\

\midrule
\multirow{3}{*}{Insert Square}
& \multirow{3}{*}{60} 
& ID  & 18.8\% 
& \textbf{100.0\% (+81.3\%)} 
& 81.3\% (+62.5\%) 
& 62.5\% (+43.8\%) \\
&                          
& OOD & 0.0\%  
& \textbf{100.0\% (+100.0\%)} 
& 25.0\% (+25.0\%) 
& 25.0\% (+25.0\%) \\
&
& Overall & 15.0\%
& \textbf{100.0\% (+85.0\%)}
& 70.0\% (+55.0\%)
& 55.0\% (+40.0\%) \\

\midrule
Fold Towel
& 100
& Overall 
& 10.0\% 
& \textbf{75.0\% (+65.0\%)} 
& 40.0\% (+30.0\%) 
& 45.0\% (+35.0\%) \\

\midrule
\rowcolor{gray!12}
Average
& 66
& Overall 
& 20.0\% 
& \textbf{90.0\% (+70.0\%)} 
& 55.0\% (+35.0\%) 
& 60.0\% (+40.0\%) \\
\bottomrule
\end{tabular}
\end{table}

Table~\ref{tab:main_results} reports the real-world adaptation performance of UniSteer and all baselines across four manipulation tasks. UniSteer consistently achieves the highest success rate on all tasks, improving the average success rate from 20\% for the initial policy to 90\% after adaptation. Compared with DSRL, which performs reinforcement learning only in the noise space without human guidance, UniSteer improves the average success rate by 35 percentage points. Compared with DAgger, which uses human corrections for supervised action-space imitation, UniSteer improves the average success rate by 30 percentage points. These results show that translating human corrections into noise-space supervision substantially improves the efficiency of noise-space policy optimization.

The improvement is consistent across tasks with different manipulation characteristics. On Pick up Spoon and Stack Blocks, UniSteer achieves 90\% and 95\% overall success rates, respectively, indicating strong adaptation performance on pick-and-place and spatial rearrangement tasks. On Insert Square, UniSteer reaches 100\% success, showing that the proposed noise-space steering is particularly effective for precision-sensitive contact-rich manipulation. Fold Towel remains the most challenging task due to deformable-object dynamics and higher execution variability, but UniSteer still improves the success rate to 75\%, outperforming both DSRL and DAgger by a large margin.

The ID/OOD split further shows that UniSteer improves not only in-distribution adaptation but also spatial generalization to unseen object placements. For all three position-based tasks, the initial policy completely fails on OOD positions, obtaining 0\% OOD success. After adaptation, UniSteer achieves 100\% OOD success on Pick up Spoon, Stack Blocks, and Insert Square. In contrast, DSRL remains weak on OOD positions, achieving 0\%, 0\%, and 25\% success on the three tasks, respectively. DAgger improves OOD performance on some tasks but is less consistent, achieving 75\%, 100\%, and 25\%. These results suggest that human-guided noise-space supervision helps the policy adapt beyond the demonstrated state distribution, while still preserving strong in-distribution performance.

Overall, these results show that UniSteer improves both sample-efficient policy adaptation and final task performance under limited real-world interaction budgets, while also enhancing spatial generalization to unseen initial object positions.

\subsection{Analysis of Adaptation Efficiency}

\begin{figure}[t]
  \centering
  \begin{subfigure}[t]{0.56\linewidth}
    \centering
    \includegraphics[width=\linewidth]{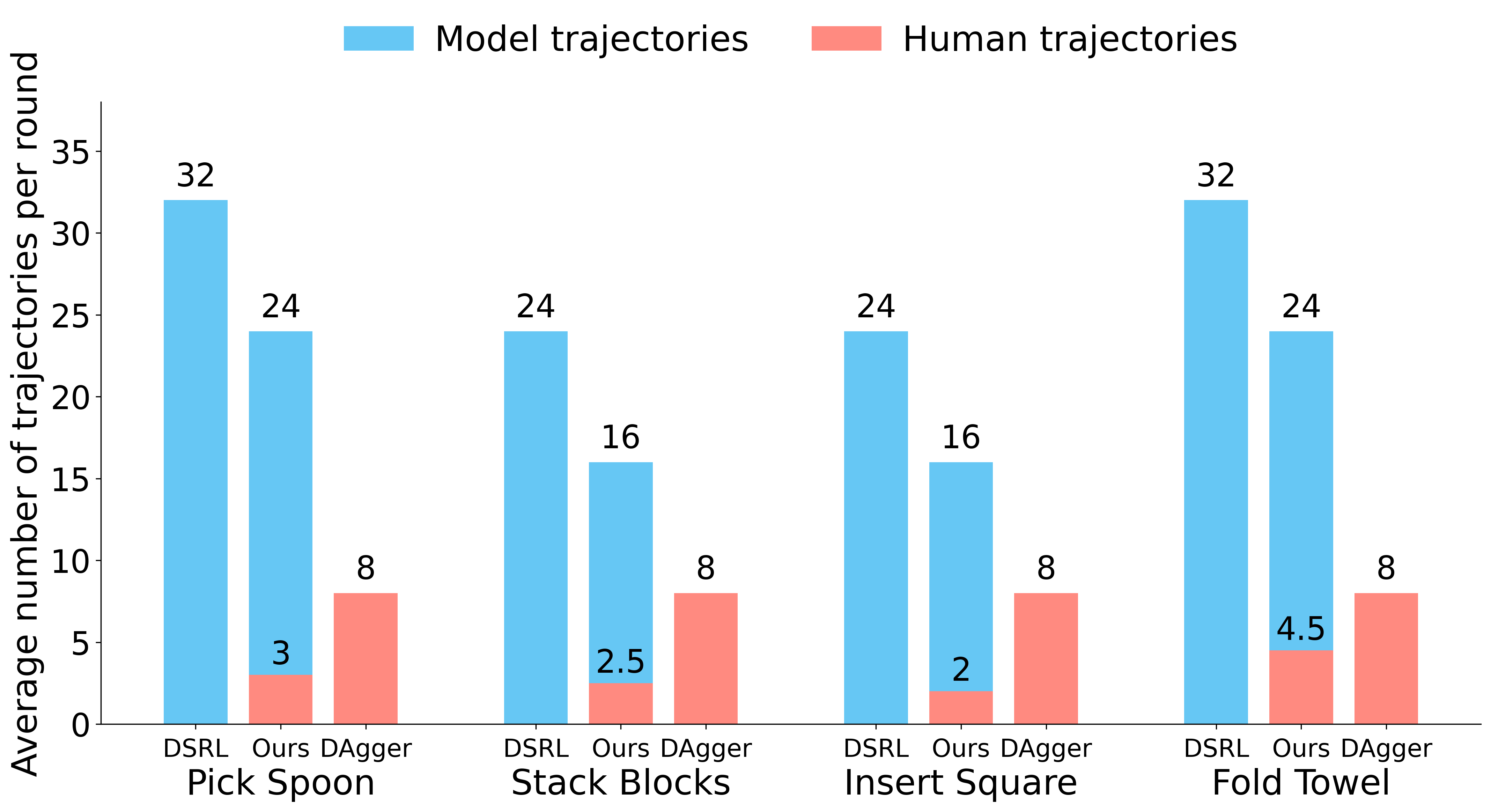}
    \caption{
     Trajectory composition per adaptation round. 
      }
    \label{fig:traj_composition}
  \end{subfigure}
  \hfill
  \begin{subfigure}[t]{0.38\linewidth}
    \centering
    \includegraphics[width=\linewidth]{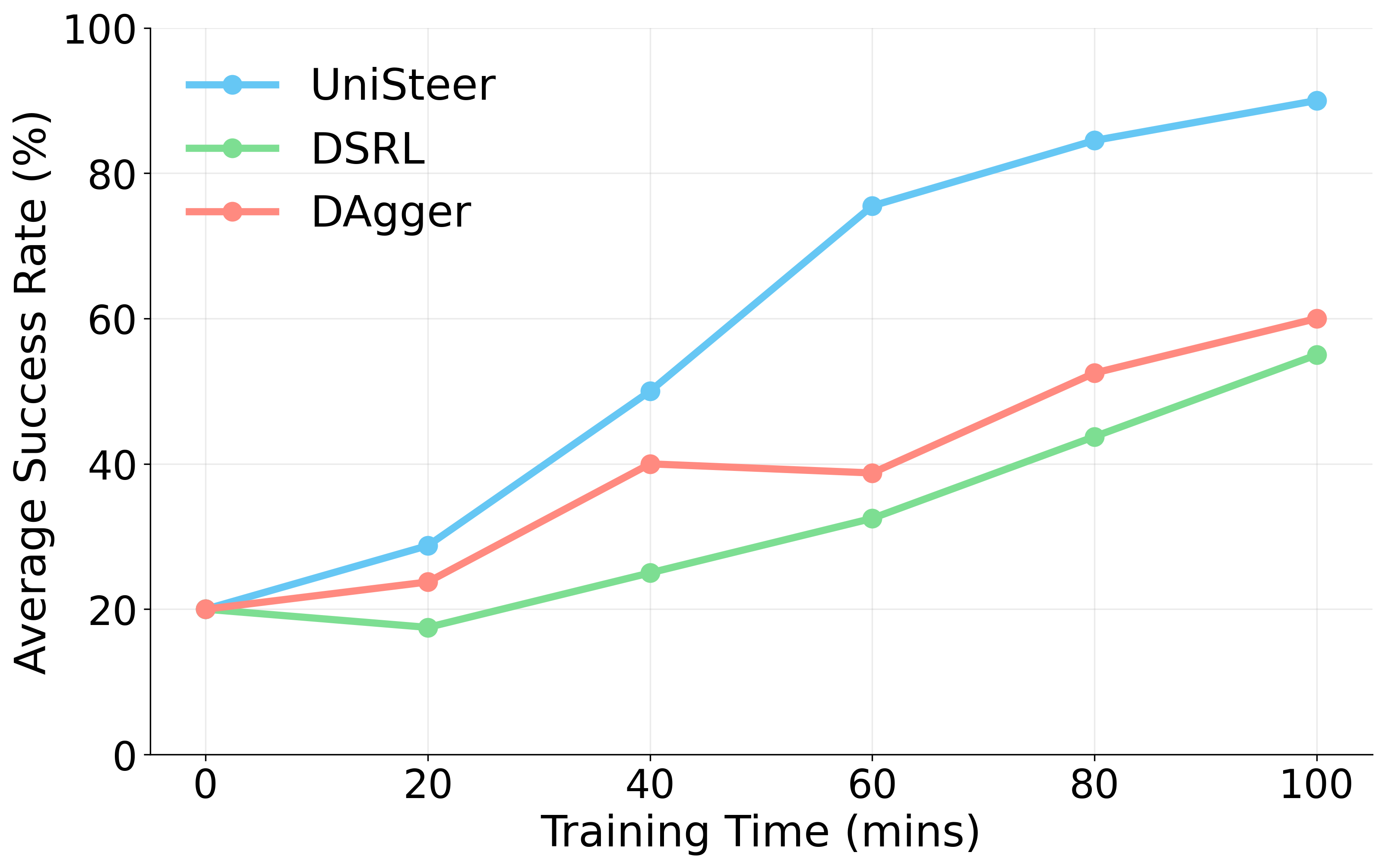}
    \caption{
     Average success rate over real-world adaptation time across four tasks.
      }
\label{fig:online_adaptation_curve}
  \end{subfigure}

  \caption{
    Online adaptation efficiency of UniSteer. UniSteer uses human guidance more efficiently to accelerate noise-space policy adaptation.
    }
  \label{fig:online_adaptation}
\end{figure}

We further analyze the trajectory composition collected during each adaptation round and present the results in Figure~\ref{fig:traj_composition}.
Compared with DSRL, which relies entirely on autonomous model rollouts, UniSteer uses fewer model-only trajectories while achieving substantially higher final success rates. This suggests that the gain of UniSteer does not come from collecting more autonomous experience, but from using human corrective signals to guide exploration more effectively. In contrast, DAgger relies exclusively on human-collected trajectories, requiring 8 pure human trajectories per round for every task. UniSteer requires far fewer pure human trajectories, averaging only 0.98 per round across tasks, while additionally leveraging a small number of mixed trajectories where the model acts autonomously before human takeover.

This trajectory composition highlights the key efficiency advantage of UniSteer. Rather than treating human interventions only as action-space demonstrations, UniSteer converts them into noise-space supervision targets and uses them jointly with reinforcement learning. As a result, even a small number of human and mixed trajectories can steer the noise actor toward more useful exploration, leading to better adaptation performance than both purely autonomous noise-space RL and action-space imitation from human corrections.

Figure~\ref{fig:online_adaptation_curve} shows the average online adaptation curve across the four real-world tasks. UniSteer improves more rapidly than both DSRL and DAgger throughout training, indicating better time-to-performance under the same real-world interaction budget. This suggests that human corrective guidance, when translated into noise-space supervision, accelerates policy improvement rather than merely improving the final converged performance.

The different learning trends also reflect the limitations of the two baselines. DSRL relies only on autonomous exploration in noise space, which makes improvement slow under sparse rewards and real-world interaction constraints. This issue is especially pronounced when the initial success rate is low: successful trajectories are rarely collected, so the policy receives limited useful reward signals and exploration becomes highly inefficient. DAgger benefits from human corrections early in training, but action-space imitation alone does not consistently translate corrections into robust policy improvement. In contrast, UniSteer uses human interventions to steer the noise actor toward more promising noise regions while continuing to optimize with reinforcement learning, leading to faster and more stable adaptation.

\subsection{Analysis of Action-to-Noise Supervision Strategies}

\begin{table}[t]
\centering
\small
\setlength{\tabcolsep}{3.5pt}
\caption{Comparison of different action-to-noise supervision strategies on 16 trajectories from Pick up Spoon and Insert Square. Inv. Time denotes the time used to recover noise targets, Train Time denotes the time used to train the noise model, Act. Loss denotes the reconstructed action-chunk MSE between the decoded action and the target human action, and SR denotes the success rate over the evaluated trajectories.}
\label{tab:supervision_ablation}
\vspace{4pt}
\begin{tabular}{llccccc}
\toprule
Task & Method & Inv. Time & Train Time & Act. Loss & Total Time & SR \\
     &        & (s)       & (s)        &           & (s)        &    \\
\midrule
\multirow{3}{*}{Pick up Spoon}
& Fixed-point inversion (Ours) & \textbf{23.05} & 50.21 & \textbf{0.00122} & \textbf{73.26} & \textbf{8/8} \\
& Optimization-based inv.      & 157.83 & 50.21 & 0.06516 & 208.04 & 3/8 \\
& Direct action supervision    & N/A    & 193.22 & N/A    & 193.22 & \textbf{8/8} \\
\midrule
\multirow{3}{*}{Insert Square}
& Fixed-point inversion (Ours) & \textbf{10.32} & 30.10 & \textbf{0.00018} & \textbf{40.42} & \textbf{8/8} \\
& Optimization-based inv.      & 50.86 & 30.10 & 0.05624 & 80.96 & 4/8 \\
& Direct action supervision    & N/A   & 90.43 & N/A     & 90.43 & \textbf{7/8} \\
\bottomrule
\end{tabular}
\end{table}

We compare fixed-point inversion with two alternative strategies for using action-space supervision. \emph{Optimization-based inversion} treats the noise as an optimizable variable and searches for a target noise that reconstructs the human corrective action through the frozen decoder. \emph{Direct action supervision} does not construct an explicit noise target; instead, it feeds the actor-predicted noise into the frozen decoder and updates the actor through action reconstruction loss.

Table~\ref{tab:supervision_ablation} shows that fixed-point inversion achieves the best trade-off among efficiency, inversion accuracy, and downstream success. Compared with optimization-based inversion, it recovers more reliable noise targets with much lower computational cost. Compared with direct action supervision, it avoids expensive backpropagation through the frozen decoder during actor updates and reduces the risk of shifting the learned noise distribution away from regions where the decoder is well calibrated. These results support our design choice of using fixed-point inversion to construct explicit, model-consistent noise targets. This provides an efficient and stable supervision signal for learning from human corrective actions in noise space.

\subsection{Analysis of Fixed-Point Inversion Iterations}

\begin{table}[t]
\centering
\small
\setlength{\tabcolsep}{6pt}
\caption{Fixed-point inversion performance under different numbers of fixed-point iterations $M$ on all human correction data. We report the average inversion time per sample and the mean, median, and 90th percentile of the reconstructed action loss, computed as the MSE between the decoded action chunk and the target human action chunk.}
\label{tab:fixed_point_iterations}
\vspace{4pt}
\begin{tabular}{ccccc}
\toprule
$M$ & Time / Sample (s) & Mean Act. Loss & Median Act. Loss & P90 Act. Loss \\
\midrule
4  & 0.0447 & 0.002526 & 0.000753 & 0.003939 \\
8  & 0.0670 & 0.002096 & 0.000224 & 0.003683 \\
16 & 0.1117 & 0.002017 & 0.000111 & 0.003631 \\
32 & 0.1980 & 0.002004 & 0.000094 & 0.003618 \\
\bottomrule
\end{tabular}
\end{table}

We analyze the effect of the number of fixed-point iterations $M$ on inversion quality and computation cost. For each recovered noise target, we decode it through the frozen flow-matching decoder and measure the reconstructed action-chunk MSE against the original human correction. Table~\ref{tab:fixed_point_iterations} shows that fixed-point inversion already achieves low reconstruction error with a small number of iterations, and that increasing $M$ yields quickly diminishing returns while steadily increasing computation. This indicates that the fixed-point iteration converges rapidly in the human-correction regime and produces model-consistent noise targets efficiently. We use $M=16$ as a practical trade-off, since it achieves nearly saturated reconstruction accuracy while remaining efficient for online adaptation.

\subsection{Analysis of Training Schedule}

\begin{table}[t]
\centering
\footnotesize
\setlength{\tabcolsep}{3.2pt}
\renewcommand{\arraystretch}{1.08}
\caption{Comparison of different training-stage schedules on Pick up Spoon and Insert Square. For each task, Overall is evaluated over 20 trials, while ID and OOD are evaluated over 16 in-distribution and 4 out-of-distribution trials, respectively.}
\label{tab:training_stage_ablation}
\vspace{4pt}
\begin{tabular}{lcrrrr}
\toprule
\multirow{2}{*}{Task} & \multirow{2}{*}{Split} & \multicolumn{4}{c}{Success Rate (\%)} \\
\cmidrule(lr){3-6}
& & SFT-then-RL & RL-then-SFT & Only SFT & Only RL \\
\midrule

\multirow{3}{*}{Pick up Spoon}
& ID      & \textbf{87.5\%} & 68.8\% & 81.3\% & 62.5\% \\
& OOD     & \textbf{100.0\%} & \textbf{100.0\%} & 75.0\% & 0.0\% \\
& Overall & \textbf{90.0\%} & 75.0\% & 80.0\% & 50.0\% \\
\midrule

\multirow{3}{*}{Insert Square}
& ID      & \textbf{100.0\%} & 81.3\% & 81.3\% & 81.3\% \\
& OOD     & \textbf{100.0\%} & \textbf{100.0\%} & \textbf{100.0\%} & 25.0\% \\
& Overall & \textbf{100.0\%} & 85.0\% & 85.0\% & 70.0\% \\
\midrule

\rowcolor{gray!12}
Average
& Overall & \textbf{95.0\%} & 80.0\% & 82.5\% & 60.0\% \\
\bottomrule
\end{tabular}
\end{table}

Table~\ref{tab:training_stage_ablation} ablates the effect of different training-stage schedules. The SFT-then-RL schedule used in UniSteer achieves the best overall performance, suggesting that the two stages play complementary roles. The supervised stage first pulls the noise actor toward target noise regions inferred from human corrections, providing a more effective starting point for exploration. The subsequent RL stage then further improves the policy through task-level rewards, leading to both more efficient exploration and stronger final performance.

When the order is reversed, the supervised update after RL can overwrite or distort the reward-optimized noise distribution learned during reinforcement learning, reducing the benefit of prior exploration. Using only SFT suffers from limitations similar to imitation-based methods: it relies solely on corrective data, can overfit to the collected correction distribution, and may forget previously learned behaviors, leading to weaker generalization. Using only RL avoids supervised overwriting, but exploration remains inefficient when the initial success rate is low and useful reward signals are sparse. Overall, these results show that supervised noise-space steering is most effective when used to guide exploration before RL, while RL is needed to consolidate and improve the policy under task-level rewards.

\section{Conclusion}

We presented \textbf{UniSteer}, the first unified noise-space finetuning framework that integrates human corrective guidance into online RL for flow-matching VLA adaptation. The key idea is to bridge the mismatch between action-space human corrections and noise-space policy optimization through approximate action-to-noise inversion, allowing both human guidance and reward-driven learning to operate through a shared noise interface while keeping the generative decoder frozen. This formulation preserves the efficiency and stability of lightweight noise-space adaptation, while addressing its reliance on inefficient autonomous exploration. More broadly, we show that human interventions can serve not only as action-space demonstrations, but also as structured noise supervision for steering frozen generative robot policies, offering a practical path toward more efficient and scalable real-world VLA adaptation.

\newpage
\bibliographystyle{unsrtnat}
\bibliography{ref}

\newpage
\appendix

\section{Theoretical Analysis and Proofs}
\subsection{Invertibility of the Continuous Flow Decoder}
\label{app:continuous_flow_invertibility}

\paragraph{Setup and statement.}
For a fixed state $s$, consider the continuous-time flow decoder
\[
\frac{d z_t}{dt}=v_\theta(z_t,t,s), \qquad t\in[0,1],
\]
and denote the induced noise-to-action map by $G_\theta(s,z_0)=z_1$. 
We show that this map is bijective under standard Lipschitz regularity.

\paragraph{Proposition A.1.}
Assume that $v_\theta(\cdot,t,s)$ is Lipschitz continuous for all $t\in[0,1]$. Then, for any fixed $s$, $G_\theta(s,\cdot)$ is bijective.

\paragraph{Proof.}
The Lipschitz assumption guarantees existence and uniqueness of the ODE solution for every initial condition. 
For surjectivity, fix any target action $a\in\mathbb{R}^d$ and solve the reverse-time ODE initialized at $a$:
\[
\begin{gathered}
\frac{d y_\tau}{d\tau}=-v_\theta(y_\tau,1-\tau,s),
\qquad y_0=a, \\
z_0:=y_1,\qquad z_t:=y_{1-t}.
\end{gathered}
\]
Then
\[
\frac{d z_t}{dt}
=
-\frac{d y_\tau}{d\tau}\Big|_{\tau=1-t}
=
v_\theta(y_{1-t},t,s)
=
v_\theta(z_t,t,s),
\qquad
z_1=y_0=a.
\]
Thus $z_t$ is a valid forward trajectory of the original ODE and $G_\theta(s,z_0)=a$, proving surjectivity. 
For injectivity, if $G_\theta(s,z_0)=G_\theta(s,z_0')=a$, then the reverse-time ODE from the same terminal value $a$ has a unique solution, so the recovered initial values must coincide: $z_0=z_0'$. 
Therefore, $G_\theta(s,\cdot)$ is bijective.
\hfill $\square$

\subsection{One-Step Fixed-Point Inversion}
\label{app:one_step_inversion}

\paragraph{Setup.}
We analyze the inverse of one Euler step of the frozen flow decoder. 
The forward step, inverse equation, and associated fixed-point map are
\[
\begin{gathered}
y=x+\Delta t\,v_\theta(x,t_k,s), \\
x=y-\Delta t\,v_\theta(x,t_k,s), \\
g_y(x):=y-\Delta t\,v_\theta(x,t_k,s).
\end{gathered}
\]

\paragraph{Proposition A.2.}
Assume that $v_\theta(\cdot,t_k,s)$ is $L$-Lipschitz and $\Delta t L<1$. Then $g_y$ is a contraction, admits a unique fixed point $x^\star$, and the iteration
\[
x^{(m+1)}=g_y(x^{(m)})
\]
converges to $x^\star$ from any initialization.

\paragraph{Proof.}
For any $x_1,x_2$, the Lipschitz constant of $g_y$ is bounded by
\[
\begin{aligned}
\|g_y(x_1)-g_y(x_2)\|
&=
\Delta t
\|v_\theta(x_1,t_k,s)-v_\theta(x_2,t_k,s)\| \\
&\le
\Delta t L\|x_1-x_2\|.
\end{aligned}
\]
Since $\Delta t L<1$, $g_y$ is a contraction. 
By the Banach fixed-point theorem, it admits a unique fixed point $x^\star$, and the Picard iteration converges to $x^\star$ from any initialization.
\hfill $\square$

\subsection{Finite-Step Approximation Error}
\label{app:finite_step_error}

\paragraph{Setup and statement.}
In practice, the exact fixed point is approximated by a finite number of fixed-point iterations. 
Let $x^\star$ be the unique fixed point of $g_y$, and let $\rho:=\Delta t L<1$. 
Then the $M$-step approximation satisfies
\[
\|x^{(M)}-x^\star\|
\le
\rho^M \|x^{(0)}-x^\star\|.
\]

\paragraph{Proof.}
The contraction property gives
\[
\begin{aligned}
\|x^{(m+1)}-x^\star\|
&=
\|g_y(x^{(m)})-g_y(x^\star)\| \\
&\le
\rho \|x^{(m)}-x^\star\|.
\end{aligned}
\]
Recursively applying this inequality yields
\[
\|x^{(M)}-x^\star\|
\le
\rho^M \|x^{(0)}-x^\star\|.
\]
An alternative a posteriori bound is
\[
\|x^{(M)}-x^\star\|
\le
\frac{\rho^M}{1-\rho}\|x^{(1)}-x^{(0)}\|.
\]
Therefore, the finite-step inversion error decays geometrically with the number of iterations.
\hfill $\square$

\subsection{Error Accumulation in Recursive Inversion}
\label{app:recursive_inversion_error}

\paragraph{Setup.}
We analyze the cumulative error induced by applying approximate inversion over $K$ Euler steps. 
Let $F_k^{-1}$ denote the exact inverse of the $k$-th Euler step and $\mathrm{Inv}_k^{(M)}$ denote its $M$-step fixed-point approximation. 
Starting from the human corrective action $a^h$, the exact and approximate recursive inversions are
\[
\begin{aligned}
&z_0^\star=a^h,
&&z_k^\star=F_k^{-1}(z_{k-1}^\star,s),
&&k=1,\dots,K, \\
&\hat z_0=a^h,
&&\hat z_k=\mathrm{Inv}_k^{(M)}(\hat z_{k-1},s),
&&k=1,\dots,K.
\end{aligned}
\]

\paragraph{Assumptions.}
For each step $k$, assume that the exact inverse is Lipschitz and the finite-step approximation error is geometrically bounded:
\[
\begin{aligned}
\|F_k^{-1}(y_1,s)-F_k^{-1}(y_2,s)\|
&\le
c_k\|y_1-y_2\|, \\
\|\mathrm{Inv}_k^{(M)}(y,s)-F_k^{-1}(y,s)\|
&\le
C_k\rho_k^M,
\qquad 0<\rho_k<1.
\end{aligned}
\]

\paragraph{Proposition A.3.}
Under the above assumptions, the recursive inversion error satisfies
\[
\|\hat z_K-z_K^\star\|
\le
\sum_{k=1}^{K}
\left(
C_k\rho_k^M
\prod_{\ell=k+1}^{K} c_\ell
\right).
\]
In particular, for any $\delta>0$, there exists $M_0$ such that for all $M\ge M_0$,
\[
\|\hat z_K-z_K^\star\|<\delta.
\]

\paragraph{Proof.}
Define the stepwise error $e_k:=\|\hat z_k-z_k^\star\|$. Since both recursions start from the same terminal action, $e_0=0$. 
For each step $k\ge1$, we have
\[
\begin{aligned}
e_k
&=
\|\mathrm{Inv}_k^{(M)}(\hat z_{k-1},s)
-
F_k^{-1}(z_{k-1}^\star,s)\| \\
&\le
\|\mathrm{Inv}_k^{(M)}(\hat z_{k-1},s)
-
F_k^{-1}(\hat z_{k-1},s)\| \\
&\quad+
\|F_k^{-1}(\hat z_{k-1},s)
-
F_k^{-1}(z_{k-1}^\star,s)\| \\
&\le
C_k\rho_k^M+c_k e_{k-1}.
\end{aligned}
\]
Here the first inequality adds and subtracts $F_k^{-1}(\hat z_{k-1},s)$, and the second inequality uses the approximation bound and Lipschitz continuity. 
Unrolling the recursion from $e_0=0$ gives
\[
e_K
\le
\sum_{k=1}^{K}
\left(
C_k\rho_k^M
\prod_{\ell=k+1}^{K} c_\ell
\right).
\]
Since $\rho_k\in(0,1)$ for every fixed $k$, each term converges to zero as $M\to\infty$. 
The right-hand side is a finite sum, so it also converges to zero. 
Therefore, for any $\delta>0$, there exists $M_0$ such that $M\ge M_0$ implies $\|\hat z_K-z_K^\star\|<\delta$.
\hfill $\square$

\section{Limitations}

UniSteer still has several limitations. First, its action-to-noise inversion depends on the frozen flow-matching decoder, and the recovered noise targets may be slightly shifted from the initial distribution of the noise actor. Second, our experiments are conducted on four tasks with one robot platform, and broader evaluation across more robots, tasks, and longer-horizon settings is needed.

\section{Algorithm of UniSteer Framework}

\begin{algorithm}[h]
\caption{Algorithm of UniSteer Framework}
\label{alg:latent_human_rl_short}
\begin{algorithmic}[1]
\Require Frozen flow matching policy $G_\theta$, noise actor $\psi_\phi$, noise-space critic $Q_\omega$, RL buffer $\mathcal{B}_{\mathrm{RL}}$, demo buffer $\mathcal{B}_{\mathrm{demo}}$
\For{each rollout}
    \For{each time step $t$}
        \State Sample noise $z_t \sim \psi_\phi(\cdot \mid s_t)$ and decode action chunk $a_t = G_\theta(s_t,z_t)$
        \If{human takeover with corrective action chunk $a_t^h$}
            \State Recover noise target $\hat z_t^h \gets \mathrm{Invert}(s_t,a_t^h)$
            \State Execute $a_t^h$ and observe $(r_t, s_{t+1})$
            \State Store $(s_t,\hat z_t^h,r_t,s_{t+1})$ in $\mathcal{B}_{\mathrm{RL}}$ and $\mathcal{B}_{\mathrm{demo}}$
        \Else
            \State Execute $a_t$ and observe $(r_t, s_{t+1})$
            \State Store $(s_t,z_t,r_t,s_{t+1})$ in $\mathcal{B}_{\mathrm{RL}}$
        \EndIf
    \EndFor

    \For{each update step}
        \State Update $\psi_\phi$ with $\mathcal{L}_{\mathrm{demo}}$ on $\mathcal{D}_{demo} \sim \mathcal{B}_{demo}$
        \State Update $Q_\omega$ , $\psi_\phi$ with $\mathcal{L}_{Q_\omega}$ , $\mathcal{L}_{RL}$ on $\mathcal{D}_{RL} \sim \mathcal{B}_{RL}$
        
    \EndFor
\EndFor
\end{algorithmic}
\end{algorithm}

\section{Hyperparameters}

\begin{table}[h]
\centering
\footnotesize
\setlength{\tabcolsep}{5pt}
\renewcommand{\arraystretch}{1.08}
\caption{Implementation hyperparameters used in our experiments.}
\label{tab:hyperparameters}
\vspace{4pt}
\begin{tabular}{ll|ll}
\toprule
Component & Value & Component & Value \\
\midrule
$\pi_0$ denoising steps & 10 
& Action horizon & 16 \\

Action dimension & 7 
& State dimension & 2055 \\

Image size & $224 \times 224 \times 6$ 
& Noise representation dim & 50 \\

Actor hidden dims & $[1024,1024,1024]$ 
& Critic hidden dims & $[1024,1024,1024]$ \\

CNN features & $[32,32,32,32]$ 
& CNN strides & $[3,2,2,2]$ \\

Replay batch size & 256 
& Replay capacity & 33333 \\

Discount $\gamma$ & 0.99 
& Target update rate $\tau$ & 0.005 \\

Actor LR & $1\times10^{-4}$ 
& Critic LR & $3\times10^{-4}$ \\

Temperature LR & $3\times10^{-4}$ 
& Actor-SFT LR & $6\times10^{-5}$ \\

Learnable temperature & True 
& Initial temperature & 1.0 \\

Target entropy & 0.0 
& $\log\sigma_{\min}$ & $-20.0$ \\

\bottomrule
\end{tabular}
\end{table}

\section{Case Study}

\begin{figure}[h]
  \centering
  \begin{subfigure}[t]{\linewidth}
    \centering
    \includegraphics[width=\linewidth]{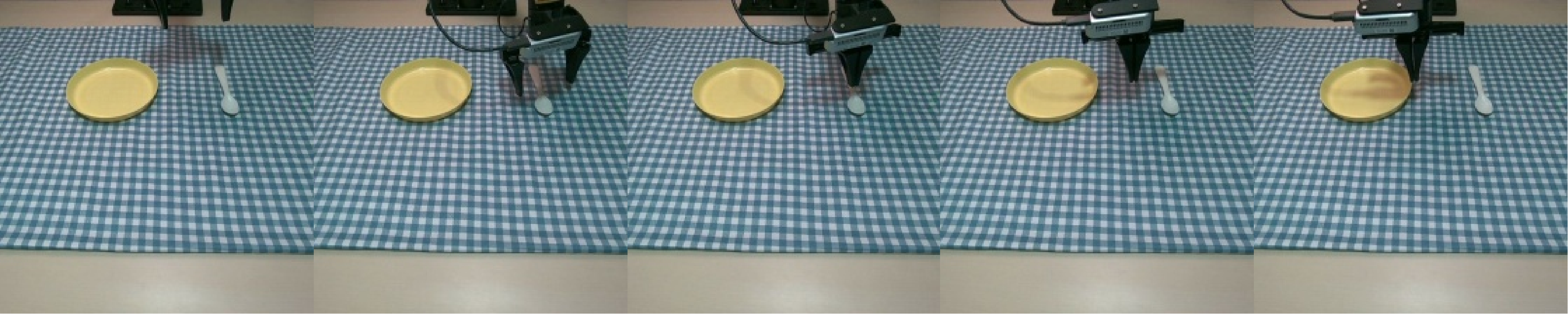}
    \caption{DSRL rollout during early adaptation.}
    \label{fig:case_study_dsrl}
  \end{subfigure}

  \vspace{0.5em}

  \begin{subfigure}[t]{\linewidth}
    \centering
    \includegraphics[width=\linewidth]{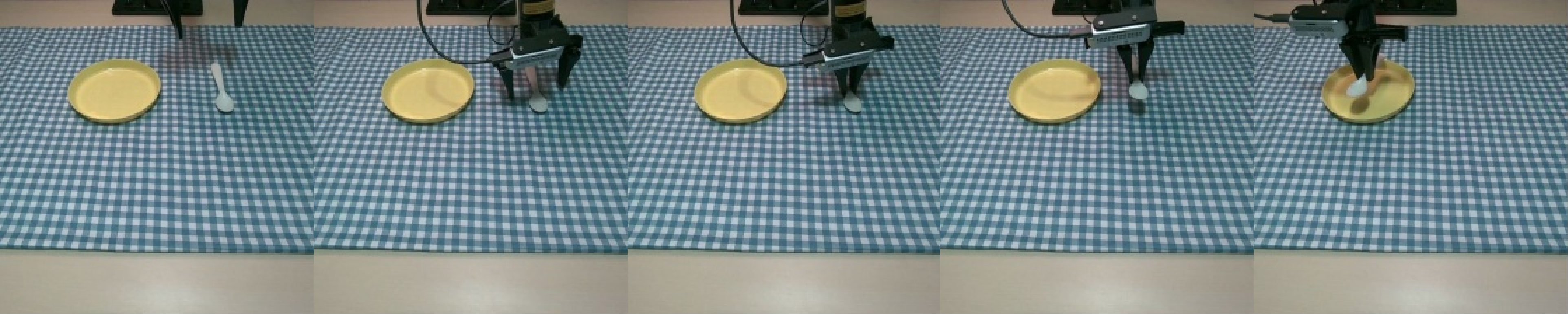}
    \caption{UniSteer rollout during early adaptation with human-guided noise supervision.}
    \label{fig:case_study_unisteer}
  \end{subfigure}

  \caption{
  Qualitative comparison of early exploration behavior on Pick up Spoon.
  DSRL relies on autonomous noise-space exploration and repeatedly approaches the spoon from above, which rarely leads to successful grasping.
  In contrast, UniSteer uses action-to-noise inversion to convert human corrective actions into noise-space supervision, guiding the noise actor toward lower and more successful grasping motions.
  }
  \label{fig:case_study_spoon}
\end{figure}

Figure~\ref{fig:case_study_spoon} provides a qualitative comparison of early exploration behavior on the Pick up Spoon task. During early adaptation, DSRL relies entirely on autonomous noise-space exploration. As shown in Figure~\ref{fig:case_study_dsrl}, the robot repeatedly approaches the spoon from above but fails to reach a graspable pose. This leads to almost no successful trajectories, making reward-driven improvement inefficient under sparse task rewards.

In contrast, UniSteer uses human corrective actions to provide noise-space supervision through action-to-noise inversion. As shown in Figure~\ref{fig:case_study_unisteer}, after receiving human-guided noise targets, the robot explores lower motions around the spoon and is able to grasp it successfully. This qualitative behavior illustrates the key advantage of UniSteer: human corrections steer the noise actor toward more promising exploration regions, enabling the policy to collect successful trajectories earlier and improving the efficiency of subsequent reinforcement learning.

\begin{figure}[h]
  \centering
  \begin{subfigure}[t]{\linewidth}
    \centering
    \includegraphics[width=\linewidth]{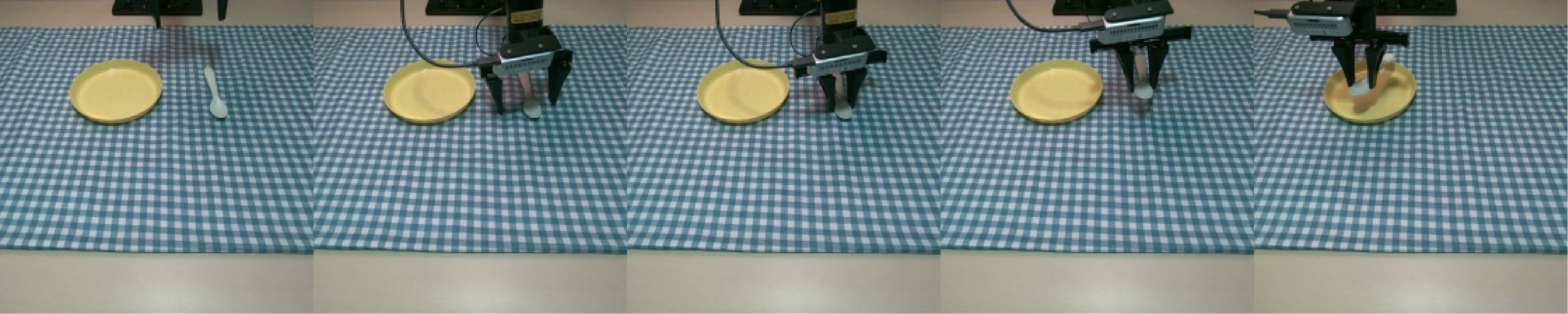}
    \caption{After the second adaptation round, DAgger learns to grasp the upright spoon.}
    \label{fig:case_study_dagger_upright}
  \end{subfigure}

  \vspace{0.5em}

  \begin{subfigure}[t]{\linewidth}
    \centering
    \includegraphics[width=\linewidth]{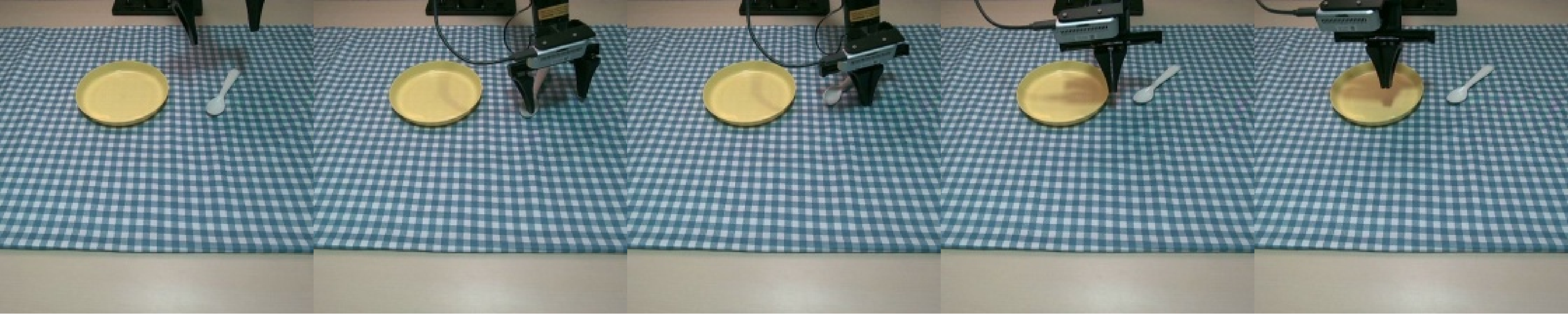}
    \caption{After the second adaptation round, DAgger forgets the tilted-spoon grasping behavior learned in the first round.}
    \label{fig:case_study_dagger_tilted_forgetting}
  \end{subfigure}

  \caption{
  Qualitative example of forgetting in DAgger on Pick up Spoon.
  DAgger learns the upright-spoon configuration in the second adaptation round, but this action-space supervised update overwrites the tilted-spoon behavior learned in the first round.
  }
  \label{fig:case_study_dagger_forgetting}
\end{figure}

Figure~\ref{fig:case_study_dagger_forgetting} illustrates a typical forgetting behavior of DAgger during online adaptation. As shown in Figure~\ref{fig:case_study_dagger_upright}, after the second adaptation round, DAgger learns to grasp the upright spoon configuration. However, when evaluated on the tilted spoon configuration that was learned in the previous round, the policy fails to preserve the earlier behavior, as shown in Figure~\ref{fig:case_study_dagger_tilted_forgetting}. This suggests that directly aggregating corrective actions in action space can introduce interference between different correction distributions, causing newly learned behaviors to overwrite earlier ones. In contrast, UniSteer translates human corrections into noise-space supervision and combines them with reward-driven optimization, which helps reduce such forgetting while improving exploration.

\section{Real-world experimental setup}

\begin{figure}[h]
  \centering
  \includegraphics[width=0.75\linewidth]{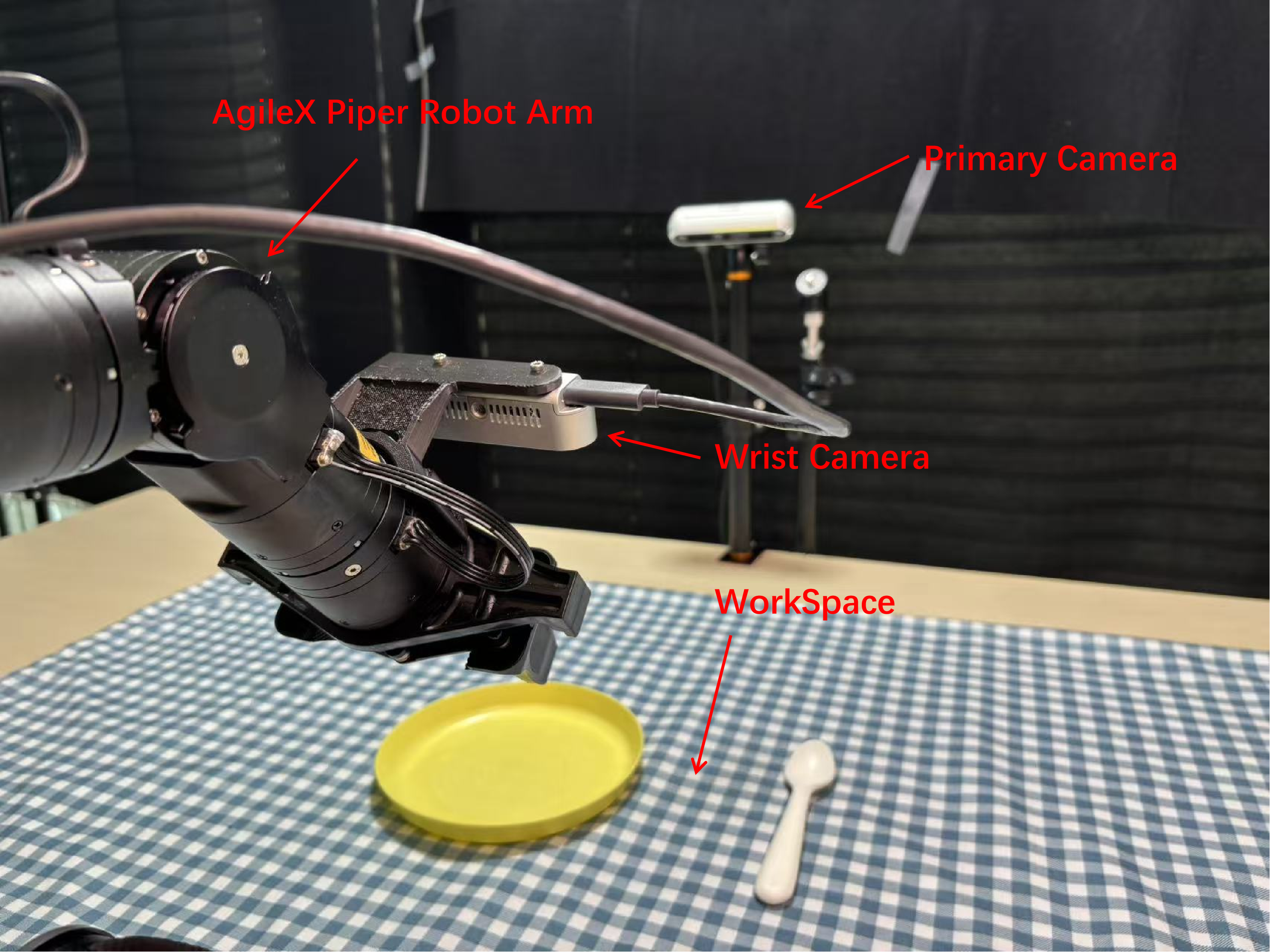}
  \caption{
  We use an AgileX Piper robot arm equipped with a wrist camera, together with a primary camera observing the tabletop workspace. 
  The policy takes both camera views, the robot end-effector pose, and gripper state as observations for real-world manipulation.
  }
  \label{fig:real_world_setup}
\end{figure}

\end{document}